\documentclass[10pt,twocolumn,letterpaper]{article}

\usepackage{iccv}      

\usepackage{amsmath,amsfonts,bm}

\def\eqref#1{equation~\ref{#1}}

\def\1{\bm{1}}

\DeclareMathAlphabet{\mathsfit}{\encodingdefault}{\sfdefault}{m}{sl}
\SetMathAlphabet{\mathsfit}{bold}{\encodingdefault}{\sfdefault}{bx}{n}

\usepackage{hyperref}
\usepackage{url}

\usepackage[utf8]{inputenc} 
\usepackage[T1]{fontenc}    
\usepackage{url}            
\usepackage{booktabs}       
\usepackage{amsfonts}       
\usepackage{nicefrac}       
\usepackage{microtype}      
\usepackage{xcolor}         
\usepackage{algpseudocode,algorithm,algorithmicx}

\usepackage{graphicx}
\usepackage{soul}
\usepackage{times}
\usepackage{epsfig}
\usepackage{dsfont}
\usepackage{amsmath}
\usepackage{amssymb}
\usepackage{tabularx}
\usepackage{makecell}
\usepackage{bm}
\usepackage{multirow} 
\usepackage{rotating}
\usepackage{wrapfig}
\usepackage{amsthm}
\usepackage{thmtools}
\usepackage{thm-restate}
\usepackage{flexisym}
\usepackage{colortbl}
\usepackage{ulem}
\usepackage{appendix}
\usepackage{ bbold }
\usepackage{caption}

\newcolumntype{C}[1]{>{\centering\arraybackslash}p{#1}}

\definecolor{adv_red}{RGB}{225,100,100}
\definecolor{adv_blue}{RGB}{60,120,215}

\graphicspath{{./images/}}

\newcommand{\emphsize}[1]{{\color{blue}#1}}

\newcommand{\xp}[1]{{\color{red}#1}}

\usepackage[symbol]{footmisc}
\renewcommand{\thefootnote}{\fnsymbol{footnote}}

    \makeatletter
\def\@fnsymbol#1{\ensuremath{\ifcase#1\or \dagger\or \ddagger\or
   \mathsection\or \mathparagraph\or \|\or **\or \dagger\dagger
   \or \ddagger\ddagger \else\@ctrerr\fi}}
    \makeatother

\title{Category-Level Multi-Part Multi-Joint 3D Shape Assembly}

\author{Yichen Li\textsuperscript{1, 2}\thanks{Work done during graduate school at Stanford.} \hspace{1mm}
Kaichun Mo\textsuperscript{3} \hspace{1mm}
Yueqi Duan\textsuperscript{4} \hspace{1mm}
He Wang\textsuperscript{5} \hspace{1mm}
Jiequan Zhang\textsuperscript{1} \hspace{1mm}
Lin Shao\textsuperscript{6} \\
Wojciech Matusik\textsuperscript{2} \hspace{1mm}
Leonidas Guibas\textsuperscript{1} \\
\textsuperscript{1}Stanford University \hspace{1mm}
\textsuperscript{2}MIT CSAIL \hspace{1mm}
\textsuperscript{3}NVIDIA \hspace{1mm}
\textsuperscript{4}Tsinghua University \hspace{1mm}
\textsuperscript{5}Peking University  \\ 
\textsuperscript{6}National University of Singapore \\ 
}

\usepackage[capitalize]{cleveref}
\crefname{section}{Sec.}{Secs.}
\Crefname{section}{Section}{Sections}
\Crefname{table}{Table}{Tables}
\crefname{table}{Tab.}{Tabs.}

\iccvfinalcopy 

\def\iccvPaperID{2429} 

\ificcvfinal\fi

\begin{document}
\maketitle

\begin{abstract}
\vspace{-2mm}
Shape assembly composes complex shapes geometries by arranging simple part geometries and has wide applications in autonomous robotic assembly and CAD modeling.
Existing works focus on geometry reasoning and neglect the actual physical assembly process of matching and fitting joints, which are the contact surfaces connecting different parts.
In this paper, we consider contacting joints for the task of multi-part assembly.
A successful joint-optimized assembly needs to satisfy the bilateral objectives of shape structure and joint alignment. 
We propose a hierarchical graph learning approach composed of two levels of graph representation learning. The part graph takes part geometries as input to build the desired shape structure. 
The joint-level graph uses part joints information and focuses on matching and aligning joints. The two kinds of information are combined to achieve the bilateral objectives.
Extensive experiments demonstrate that our method outperforms previous methods, achieving better shape structure and higher joint alignment accuracy. 
\end{abstract}
\vspace{-8mm}
\vspace{-1mm}
\section{Introduction}
\label{sec:intro}
Shape assembly composes complex shape geometries by arranging a set of simple or primitive part geometries. Many important tasks and applications rely on shape assembly algorithms. For example, assembling Ikea furniture requires one to identify, reorient, and connect the relevant parts. Computer-Aided Design (CAD) modeling requires designers to reposition and align a set of part geometries to create complex designs. An accurate and robust shape assembly algorithm is critical to the development of autonomous systems for furniture assembly or CAD modeling~\cite{wang2021sota, marconi2019feasibility, Suarez-Ruizeaat:2018}.

\begin{figure}
    \hspace{-1mm}
    \vspace{-0.4cm}
    \includegraphics[width=1\linewidth]{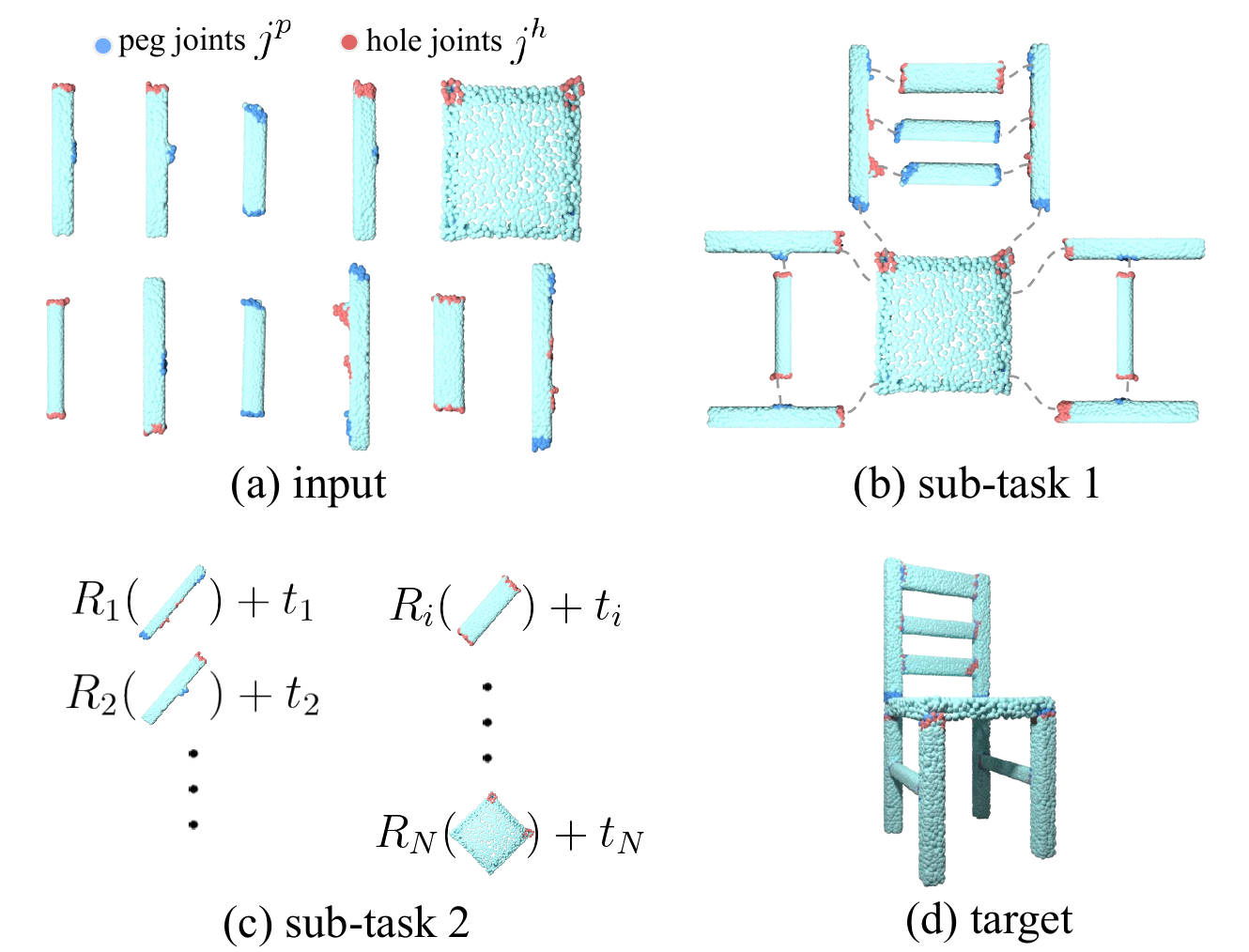}
  \vspace{-0.4cm}
      \caption{   \small{(a) Joint-annotated part point cloud input where \emphsize{blue} points indicate peg joints and \xp{red} points denote hole joints. (b) joint pairing process to produce a bipartite matching between pegs and holes (c) joint-matching-aware SE(3) pose prediction 
     (d) assembling a shape of a valid structure and with pegs and holes aligned. }}
    \label{fig:teaser}  
    \vspace{-0.5cm}
\end{figure}

\label{sec:ps}
In this paper, we aim to tackle the task of \textit{multi-part} \textit{multi-joint} shape assembly. This task simulates the real-world furniture assembly setting, where multiple shape parts are connected in different ways through contacting joints to make a complex shape geometry~\cite{lee2019ikea, lau2011converting}. As Fig.\ref{fig:teaser} shows, we are given (a) \textit{multiple shape parts}, where each part contains \textit{multiple joints}.  For our setting, we use peg-hole joint pairs to represent the allowed connections, similar to bolts and nuts where matching is only allowed between male and female pieces of the same contacting geometry~\cite{willis2022joinable}. 
Our goal is to (b) correctly connect all peg joints with hole joints and (c) piece these parts together to (d) make a desired shape. 

Many research efforts have been made towards devising shape assembly algorithms~\cite{li2020learning, qingnan, sung2017complementme, schor2019componet, willis2022joinable, NSM, Chaudhuri:2011, chaudhuri2020learning, harish2021rgl, jones2020shapeassembly, Li:2020, thomas2018learning, yanRPMNet19, yin2020coalesce, Wu:2019:pqnet, wu2021deepcad}. These prior works take the holistic geometric perspective of modeling shapes from parts. 
They produce shapes with great aesthetic value. However, this pure geometric perspective is agnostic of the rotational and reflective symmetry of parts and thus results in upside-down, flipped, and rotated part pose predictions. These noisy predictions can lead to unmatched joints or mismatches between joints, making it difficult to directly employ them in the context of autonomous assembly~\cite{Suarez-Ruizeaat:2018, xu2021zone, jimenez2013survey, willis2020fusion, thomas2018learning} or modeling of functional shapes~\cite{lin2017recovering, zheng2013smart}.
Many challenges exist in the multi-part multi-joint assembly setting: 1) large matching search space, 2) non-continuous optimization, and 3) error compounding.

Previously, Willis et al.~\cite{willis2022joinable} have considered joints for shape assembly, but they focus on assembling shapes from only \textit{two parts} with one pair of joints. In the two-part assembly setting, the pairing of joints is explicit and thus the desired assembly can be directly achieved through continuous pose optimization. 
However, our multi-part multi-joint task requires solving for a bipartite joint pairing in a very large matching space.
Additionally, our task requires interleaved \textit{discrete} and \textit{continuous} optimization. 
Joint pairing is a combinatorial problem in a discrete solution space, whereas pose estimation is in a continuous solution space. 
In the multi-part multi-joint setting, part poses need to simultaneously satisfy both the combinatorial and the continuous constraints. 
Finally, optimization of this task is sensitive to error compounding.
When one pair of joints are mismatched, the poses for the two parts need to be falsely adjusted in order to align these wrongly matched joints. These erroneous pose predictions reciprocally affect other joints on the two parts. These local errors propagate and eventually lead to the deterioration of the entire shape structure. 

To tackle these challenges, we propose an end-to-end graph learning approach in a divide-and-conquer manner. We decouple the complex task objective into combinatorial and continuous subgoals modeled by two levels of graph representation learning. The joint-level graph uses joint information and focuses on matching part joints, and the part graph takes part geometries as input to build the desired shape structure. 
The two levels of graphs are then combined to achieve both two objectives through hierarchical feature aggregation. 
Since joints are special locations on parts, we aggregate all joint-level features for each part to form a set of joint-centric part features.
These joint-centric part features are combined with the learned part graph to predict part poses to meet both the shape structure and joint matching objectives.
To alleviate the error-compounding issue, we use several graph iterations to assemble shapes in a coarse-to-fine manner. 
Each graph iteration learns to correct and refine part pose predictions from the previous iteration to eventually achieve the multi-level objectives. 
Extensive experiments demonstrate that we are able to achieve higher joint matching accuracy and more reliable shape structure over prior works. 
Our contributions are summarized as follows: 
\begin{itemize}
\vspace{-2mm}
\item We consider the concept of joint for the problem of category-level multi-part 3D shape assembly. We introduce a joint-annotated part dataset as well as a set of evaluation metrics to examine the performance.
\vspace{-2mm}
\item We propose a novel hierarchical graph network that simultaneously optimizes for both holistic shape structure and joint alignment accuracy.
\vspace{-2mm}
\item We conduct extensive experiments to demonstrate the advantages of our approach over prior works on both task objectives of holistic shape structure and joint alignment accuracy. 
\end{itemize}

\section{Related Work}
\label{sec:related}
\textit{Assembly-based 3D Modeling.} Part assembly plays an important role in many tasks~\cite{wang2019shape2motion, zheng2013smart, zakka2019form2fit, lupinetti2019content,  Lee:2019, hu2017learning}. As a pioneering work, \cite{funkhouser2004modeling} proposes a data-driven synthesis approach for 3D geometric surface model reconstruction. 
Since then, various methods have been proposed to generate shapes from parts~\cite{jones2020shapeassembly, koch2019abc, Huang:2015, han2020compositionally, gao2019sdm, gadelha2020learning, huang2006reassembling, zhang20153d}. 
Many works~\cite{chaudhuri2011probabilistic,kalogerakis2012probabilistic,jaiswal2016assembly} focus on using probabilistic graphical models to encode semantic and geometric relations among shape parts. Other works~\cite{chaudhuri2010data,shen2012structure,sung2017complementme, wu2019pq} build 3D shapes conditioned on partial shapes. While most of these methods require a third-party shape repository, some generative methods have been presented in recent years~\cite{gadelha2020learning, gao2019sdm, qingnan, jones2020shapeassembly, Li:2020, Mo:2019:Structurenet, wu2019pq}. For example, \cite{wu2019pq} first learn to generate shape parts and then estimate the transformation of parts to compose shapes. 
\cite{qingnan, harish2021rgl} design a dynamic graph learning approach by reasoning about part poses and relations iteratively. 
Our task is different from these prior assembly-based shape synthesis works in that our goal is not to generate a variety of shapes, but to solve for the set of part poses that make \textit{one} desired shape and match all the joints at the correct locations. 
Previously, \cite{li2020learning} explores the problem of single-image-guided 3D part assembly, using an image as guidance to predict 6D part poses to assemble the desired shape, but they neglect the information of part joints and contact surfaces.
Recently, Willis et al.~\cite{willis2022joinable} also considers joints for shape assembly but with only {two parts} and {one pair of joints}, whereas our task considers {multiple shape parts} and each comes with {multiple contact joints}. \cite{willis2022joinable} also assumes watertight part geometry, whereas we relax this assumption and use simple point cloud representation that can be easily obtained using commercial scanners~\cite{lin2017recovering}.

\begin{figure*}

    \vspace{-0.3cm}
    \includegraphics[width=1\linewidth]{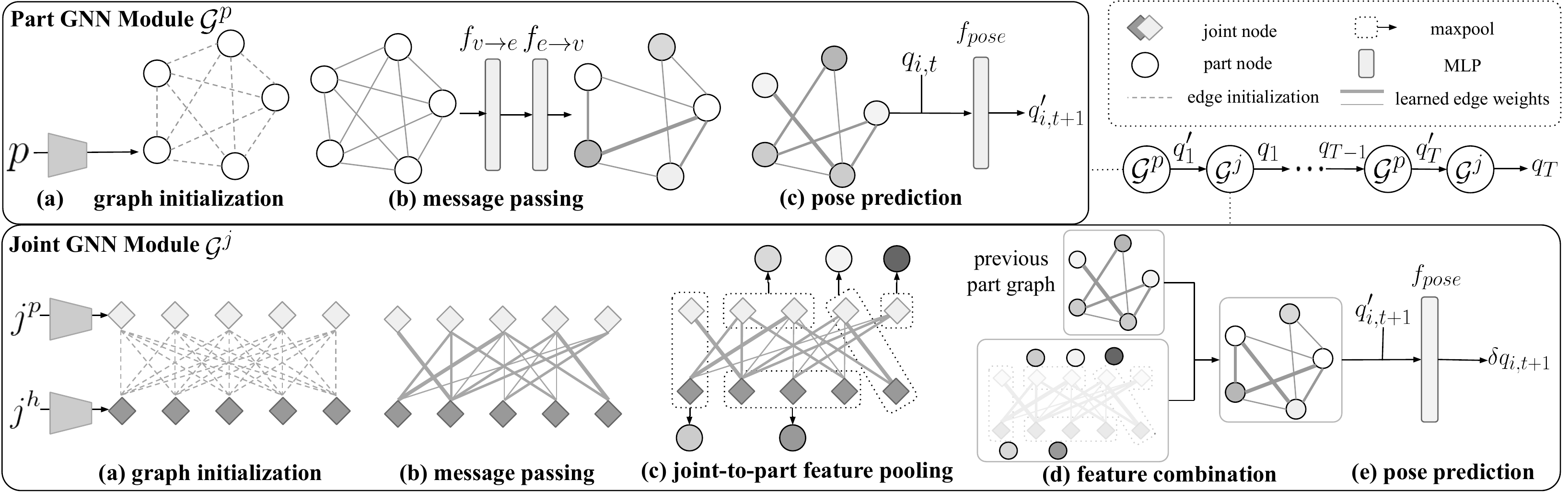}

    \caption{\small{Our multistage graph network is composed of two main GNN modules: the part graph module and the joint graph module.
    The part graph module is responsible for predicting poses for each part to construct the desired shape structure. The joint graph module helps to correct part poses to connect the matched joints. Joint graph message passing (b) contains four message-passing layers: $f_{v\rightarrow e}, f_{e \rightarrow v}, f_{v \rightarrow r}, f_{r \rightarrow v}$.}}
    \label{fig:joint_graph}
    \vspace{-0.3cm}
\end{figure*}

\textit{Graph Learning for Part Relationship} Graph neural network has been proposed to study the relationship between entities to better understand objects and scenes. Recently, a line of research \cite{xu2017scene,li2017scene,yang2018graph,chen2019scene} learns the scene graph information from the labeled object relationship
with a graph neural network, which benefits object detection on large-scale image datasets such as Visual Genome \cite{krishna2017visual}. Other works \cite{li2019grains,zhou2019scenegraphnet,wang2019planit,ritchie2019fast} explore the physical relationships between objects with geometrical and statistical heuristics, which are encoded in the iterative neural encoding and achieve decent performance on the 3D scene generation task. Inspired by the success of graph learning in various tasks, other research \cite{gao2019sdm,wu2019sagnet, qingnan, harish2021rgl, li2017grass} apply the part relation reasoning to learn shape structure and geometry for 3D shape modeling. 
These prior works deal with more apparent object-level or part-level relationships and leverage explicit relationship supervision, which is calculated from shape topology, adjacency, and support. We are different from these works in that our task deals with a hierarchy of relationships, the relationships among joints, and the relationship among parts. We use a hierarchical graph learning technique to simultaneously achieve bilateral objectives.
	
\section{Method}

\textit{Problem Setup.}
Our multi-part multi-joint shape assembly task is defined as follows:
given 1) a set of 3D part point clouds $\mathcal{P} = \{p_i\}_{i=1}^N$ and 2) each part should contain a number of peg and/or hole joints 
$\mathcal{J} = \{{j_k^{p}}, {j_k^{h}}\}_{k=1}^M$, we aim to predict a set of 6-DoF part pose $q_i=(R_i, t_i), q_i\in SE(3)$ for \textit{all} input parts $\mathcal{P}$ to satisfy the \textit{bilateral} objectives: 1) the union of the transformed parts $S=\cup_i q_i(p_i)$ that forms a desired 3D shape,  2) all joints are matched, and the matched pegs $J^p$ and holes $J^h$ are close to each other.

\label{sec:method}
\textit{Overview. } Our multi-part multi-joint shape assembly task has several challenges 1) find the set of one-to-one peg-hole matching from a very \textit{large matching search space} ($O(M^2)$), 2) predict poses for all parts such that they {simultaneously} achieve two objectives of \textit{connecting all matched joints} and \textit{forming desired shape structure}, 3) local joint matching or pose prediction errors can easily propagate to the entire shape and leads to degeneration.

To deal with the first challenge, we introduce a shape-prior heuristic to reduce the matching search space. Inspired by previous works~\cite{li2020learning, qingnan}, we use the part geometry information to propose an initial rough shape structure via a part graph. Then, our joint graph works with the rough shape structure to find an initial peg-hole matching. 
We address the second challenge by having the two levels of graph representation learning focus on each of the two objectives. The joint graph module matches joints. Part graph constructs shape. We then combine the joint-level and part-level information using hierarchical feature aggregation to predict part poses subject to both objectives. We gradually refine the part poses by alternating between the two representations to achieve the two desired objectives.

\begin{figure*}
    \centering
    \vspace{-0.3cm}
    \includegraphics[width=1\linewidth]{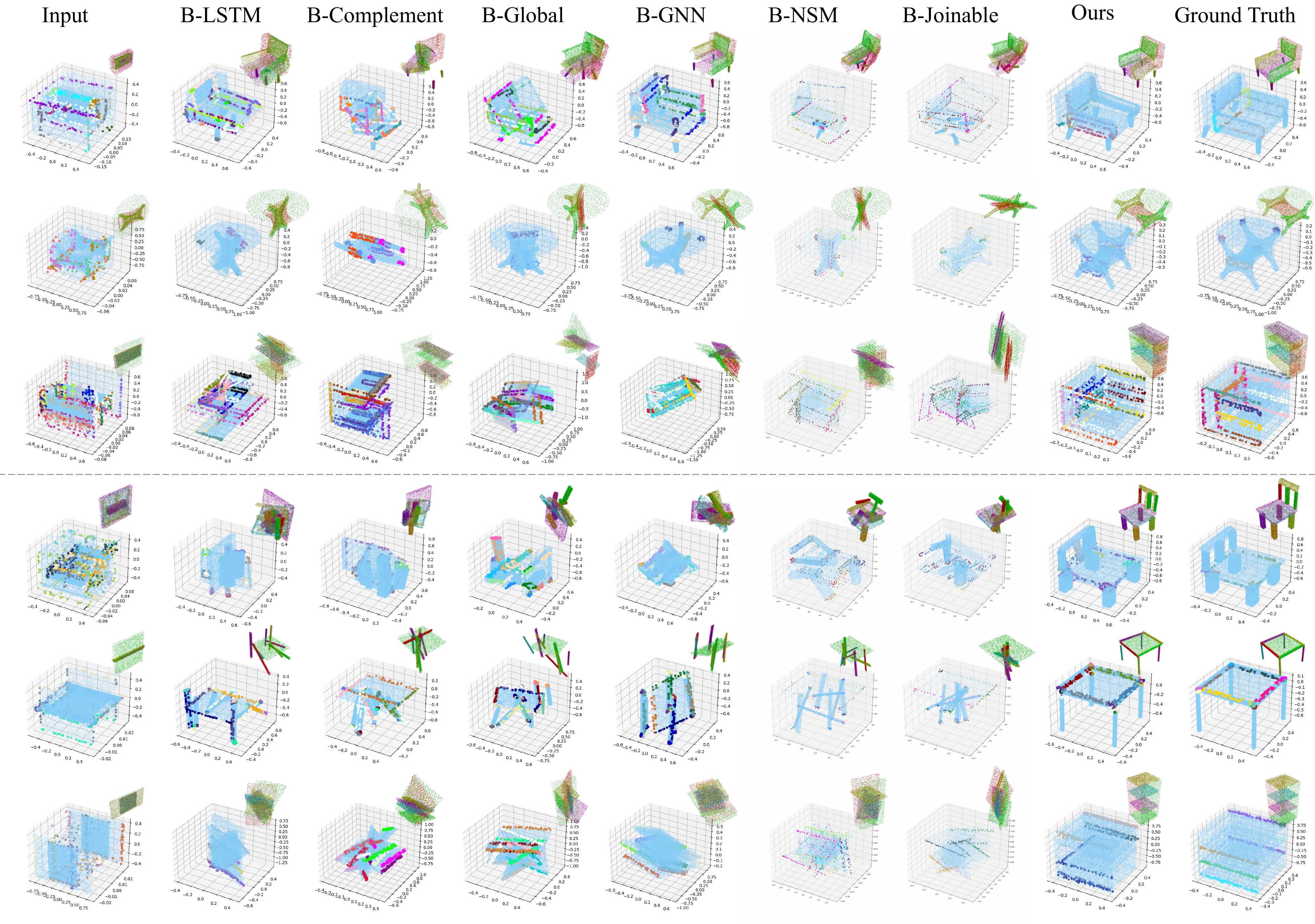}
      \caption{\small{Qualitative comparison of our method and baselines (best viewed in color). We show the predictions in both the shape view (top-right corner) and the joint view (blue shapes), where the paired joint point sets are in the same color}. In top three rows, the baselines are trained with their original setup. In bottom three rows, the baseline methods are trained with our joint input and loss. Directly imposing our proposed input and loss setting on baseline methods leads to collapsed shape prediction, whereas our proposed method produces the most structurally sound and joint optimized predictions.}
    \label{fig:qual}  
    \vspace{-0.3cm}
\end{figure*} 

\textit{Part Graph Pose Proposal.}
The part graph aims to propose desired shape structure from a given set of part geometries. Inspired by~\cite{li2020learning, qingnan}, we directly regress part poses from part geometries. Therefore, we initialize our part graph $\mathcal{G}^{p} = (\mathcal{V}^{p}, \mathcal{E}^{p})$ by encoding part geometry $\mathcal{P}$ features on each graph node $v_i^{p}$ and edges $e_{i,j}^{p}$ running between all part nodes. 
The part geometric features are extracted using PointNet~\cite{qi2017pointnet}.
In order to explicitly model the relationship between parts to form the desired shape, we use graph message passing, a mechanism for nodes to exchange information with their neighbors through edge connections.
Part-level message passing is achieved through iteratively updating the edge features ${e_{ij}^{p}} = {f_{v\rightarrow e}^{p}}({v_{i}^{p}},{v_{j}^{p}})$ and node features ${{v_{i}^{p}}'} = {f_{e \rightarrow v}^{p}}{(v_{i}, \frac{1}{N} \sum_{j=1}^{N}e_{ij})}$. We use the updated graph for pose prediction, as shown in the top section in Fig.~\ref{fig:joint_graph}. 
In the first iteration of part graph convolution, part pose vectors are decoded from the part node features, $q_{i} = f_{pose}({v_{i}^{p}}')$. For any subsequent iterations, the pose vector $q_{i, t+1}$ is predicted given the previous step pose prediction and the updated node features ${q'}_{i, t+1} = f_{pose}({v_{i}^{p}}', q_{i, t})$.

\textit{Joint Graph Relationship Reasoning. }
\label{sec:jointgraph}
We use a joint graph to infer and refine joint connectivity relationships. 
As shown in the bottom section of Fig.~\ref{fig:joint_graph}, we first initialize joint node features $v_i^{j}$ using PointNet to extract the joint-geometry feature vectors. 
The joint edges $e_{ij}$ are initialized to be a set of bipartite edges running between all pegs nodes and all holes nodes $\mathcal{E}^{j}= \{e^{j}_{p,h}\}$ to reflect all possible allowed connections. 
We then use message passing to update the edge and node features iteratively. 
Specifically, we first update features of each edge $e_{ij}$ with the neural messages calculated from its connected node features, $e_{ij} = f_{v \rightarrow e}(v_i,v_j)$. For the subsequent step, we update the node features $v_i$ by aggregating information from all connected joint edges $v_i' = f_{e \rightarrow v}(v_i, \frac{1}{M} \sum_{j=1}^Me_{ij}) $. 
	\begin{table*}[h]
        \small 
		\begin{center}
	\begin{tabular}{c|c|cccc|cccc}
				\toprule
				& & \multicolumn{4}{c}{Shape Chamfer Distance $\downarrow$} & \multicolumn{4}{c}{Part Pose Accuracy $\uparrow$} \\ 
				\midrule
				Setting& Method & Chair & Table & Cabinet & Average& Chair & Table & Cabinet & Average\\
				 \midrule				
				 \multirow{5}{*}{Original Setting}
				&B-Global &  0.015  &  0.013  &  0.008  & 0.013 &  32.8  &  30.1  &  33.6  & 31.4 \\
                &B-LSTM  &  0.017  &  0.026  &  0.007  & 0.021 &  39.4  &  22.5  &  44.4  & 30.8 \\
                &B-Complement  &  0.028  &  0.034  &  0.222  & 0.046 &  11.0  &  5.33  &  0.0  & 7.2 \\
                &B-GNN   &  0.007  &  0.008  &  0.006  & 0.007 &  65.3  &  61.4  &  45.0  & 61.7 \\
                &B-NSM   &0.013 & 0.022 & 0.012&0.018  &25.3 & 48.2 &18.9 & 37.0\\
				\midrule
				\multirow{6}{*}{\shortstack{Our Full Setting \\ (joint input and loss)}}
				&B-Global  &  0.029  &  0.022  &  0.013  & 0.024 &  5.4  &  12.0  &  15.0 & 9.6 \\
                & B-LSTM  &  0.037  &  0.029  &  0.017  & 0.031 &  4.4  &  4.1  &  15.3  & 5.1 \\
                &B-Complement  & 0.048  &  0.044  &  0.029  & 0.044 &  4.5  &  8.0  &  11.6  & 6.9 \\
                &B-GNN   &  0.034  &  0.039  &  0.021  & 0.036 &  11.5  &  3.2  &  10.4  & 7.0 \\
                &B-NSM   & 0.014& 0.032 &0.020 & 0.024 & 19.0& 12.1& 14.7 & 15.0\\
                &B-Joinable   & 0.026 & 0.037 & 0.025& 0.032 & 12.6 & 7.3 & 12.1 & 9.7 \\
			    \midrule
				&\textbf{Ours}  &  \textbf{0.006}  & \textbf{ 0.007 } & \textbf{ 0.005}  & \textbf{0.006} &  \textbf{72.8}  &  \textbf{67.4 } & \textbf{ 63.3} & \textbf{69.2} \\
				\bottomrule
			\end{tabular}
		\end{center}
        \vspace{-3mm}
		\caption{\small{Quantitative comparison for the \textbf{Shape Structure Metrics} between our approach and the baseline methods under two settings. Original Setting on the top rows shows the performance of the baselines method with inputs and losses as originally proposed. Our full setting on the bottom rows shows the baseline performance with our joint-annotated part inputs and our joint-aware losses, same to our method. Down arrow indicates that lower numerical values corresponds to better performances. Up arrow means higher number is better. }}
        \vspace{-2.5mm}
		\label{table:exp-shape}
	\end{table*}	
We further update the joint node features $v_i$ by explicitly modeling the joint connectivity relationships. 
Joint connectivity depends on two critical pieces of information, contact surface geometry, and relative part positions.
Therefore, we model the joint matching relationship from joint geometry $p^{joint}_i$ and part position $q_i$. We learn a joint connectivity matrix $r_{ij} \in [0,1]$ to reflect how joints are connected. The connectivity matrix is then used as edge weights applied to edge features $e_{ij}$, and
we further update the joint nodes by aggregating the weighted edge features. 
\begin{align}
r_{ij}^{(t)} &= f_{v \rightarrow r}\left(f_{r} \left(q_i;p^{joint}_i\right),f_{r} \left(q_j;p^{joint}_j\right)\right) \\
e_{ij}' &= e_{ij}r_{ij}, \quad v_i''  = f_{r \rightarrow v}\left(v_i',\frac{\sum_{j}e_{ij}'}{\sum_{j}r_{ij}} \right).
\end{align}

\textit{Joint-Aware Pose Prediction. }
In order to generate part poses that simultaneously achieve both joint matching and shape structure objectives, we need to combine information from both the part graph and the joint graph. 
The joint-part relationship is hierarchical since joints are the contacting locations on parts. 
We propose to model this relationship using hierarchical feature aggregation. 
Specifically, we use pooling operations on all relevant joint nodes for a part $\{{v}^{j}_{t}\}_{k=1}^{n_{i}}$ to form a new joint-centric part feature $v^{p*}_{i}$, as shown in section (c) in the bottom of Figure~\ref{fig:joint_graph}. 
\begin{equation}
    {v}_{i}^{p*} = MaxPool( \{ {v}_{k}^{j} \}_{k=1}^{n_{i}}), 
\end{equation}
These joint-aggregated part node features are then combined with the original part graph through part-wise feature concatenation for joint-aware pose prediction ${v}_{i}^{p'} = \{ {v}_{i}^{p} ; {v}_{i}^{p*} \}$, as shown in section (d) in Fig.~\ref{fig:joint_graph}. 
Now with the new part features ${v}_{i}^{p'}$ containing both joint and part information, we conduct the joint-aware pose proposal with the new updated part graph. 
Conditioning on part poses $\{q_{i, t+1}\}$ generated by previous graph iteration, we predict a refinement part pose operator $\delta q_{i, t+1} = f_{joint-pose}( {v}_{i}^{p'} | q_{i, t})$, as shown in section (e) in Fig.~\ref{fig:joint_graph}.
The new pose prediction is composed of the predicted pose operator and the previous stage part pose, 
 \begin{equation}
q_{i,t+1} = [\delta R_{i, t+1} \cdot R_{i, t+1}', \delta t_{i, t+1} + t_{i, t+1}']
 \end{equation}
where the new rotation is calculated by applying the new rotation difference on the previous rotation prediction, and translation is updated by adding the translation difference and previous translation prediction, more details in Appendix~\ref{supp:detail}.

\textit{Loss Functions. }
We leverage two sets of loss functions: shape loss $\mathcal{L}_{shape}$ and joint loss $\mathcal{L}_{joint}$ to optimize our multistage graph network. 
$\mathcal{L}_{shape}$ aims to help the part graph network to generate a valid shape structure, and $\mathcal{L}_{joint}$ helps the joint graph to match and connect all joints. 

\textbf{Shape loss}
We focus on the aspects of translation, rotation, and holistic shape structure in devising our shape loss $\mathcal{L}_{shape}$, where $\mathcal{L}_{shape} = \lambda_1 \mathcal{L}_t  + \lambda_2 \mathcal{L}_r + \lambda_3 \mathcal{L}_a$.
	We use $\mathcal{L}_2$ loss to supervise translation, and CD to supervise rotation and holistic shape structure. 
	\begin{equation}
	\begin{small}
	\begin{aligned}
\mathcal{L}_t &= \sum_{i=1}^N ||t_i - t_i^{gt}||_2^2 \text{,\hspace{4mm}} \\ 
\mathcal{L}_r &= \sum_{i=1}^N d_{chamfer}(R_i(p_i), R_i^{gt}(p_i))  \\
\mathcal{L}_a &= d_{chamfer}\left( \sum_{i=1}^N \left( q_{i}(p_i)\right), \sum_{i=1}^N\left( q_{i}^{gt}(p_i)\right)\right)
	\end{aligned}
	\end{small}
	\end{equation}
	where Chamfer Distance (CD) is defined as~\cite{achlioptas2017latent_pc}:
\begin{equation}
    d_{chamfer} \left( a, b\right) =  \sum_{x \in a} \min_{y \in b}||x-y||_2^2 + \sum_{x \in b} \min_{y \in a}||x-y||_2^2.
\end{equation}\label{equation:chamfer}
Additionally, as inspired by \cite{li2020learning}, we ensure our shape loss to be an order invariant loss metric to address the geometrically congruent parts, e.g. legs of a chair. Specifically, we perform Hungarian matching~\cite{kuhn1955hungarian} within each congruent part class to supervise with the closest ground truth part pose. 
	\begin{table*}[h]
        \small 
		\begin{center}
	\begin{tabular}{c|c|cccc|cccc}
				\toprule
				& & \multicolumn{4}{c}{Joint Chamfer Distance $\downarrow$} & \multicolumn{4}{c}{Joint Matching Accuracy $\uparrow$} \\ 
				\midrule
				Setting& Method & Chair & Table & Cabinet & Average & Chair & Table & Cabinet & Average \\
				 \midrule				
				 \multirow{5}{*}{Original Setting}
				&B-Global &   0.712  &  0.847  &  0.667  & 0.780 &  13.4  &  15.8  &  10.7  & 14.5 \\
                &B-LSTM  &   0.756  &  0.728  &  0.651  & 0.733 &  17.0  &  13.2  &  14.8  & 14.8 \\
                &B-Complement  &   0.901  &  0.977  &  1.074  & 0.954 &  7.5  &  8.3  &  23.6  & 9.2 \\
                &  B-Dynamic    &  0.725  &  0.855  &  0.683  & 0.791 &  24.4  &  30.0  &  18.6  & 26.9 \\
                &B-NSM   & 0.697 & 0.717 & 0.700 & 0.708 & 15.1 & 16.9 & 17.1 & 16.2\\
				\midrule
				\multirow{6}{*}{\shortstack{Our Full Setting \\ (joint input and loss)}}
				&B-Global   &  0.513  &  1.268  &  0.488  & 0.912 &  12.7  &  4.0  &  6.9 & 7.6 \\
                & B-LSTM  &  0.394  &  0.875  &  0.467  & 0.655 &  20.3  &  7.7  &  13.8   & 13.1 \\
                &B-Complement  &  0.456  &  0.647  &  0.503  & 0.561 &  17.2  &  15.5  &  17.0 & 16.3 \\
                &B-Dynamic  &  0.379  &  0.786  & \textbf{0.416}  & 0.598 &  21.5  &  10.3  &  20.0   & 15.4 \\
                &B-NSM   & 0.556 & 0.698 & 0.517 & 0.629 & 18.9& 12.1& 7.8 & 14.4 \\
                &B-Joinable   & 0.653 & 0.812 & 0.483 & 0.725 & 16.1 &13.9 & 9.4& 14.4\\
			    \midrule

&\textbf{Ours}  &  \textbf{0.352}  & \textbf{ 0.602 } &  0.620  & \textbf{0.505} & \textbf{ 57.2}  &  \textbf{50.6}  &  \textbf{27.5}  & \textbf{51.4} \\
				\bottomrule
			\end{tabular}
		\end{center}
        \vspace{-3.5mm}
		\caption{\small{Quantitative comparison for the \textbf{Joint Matching Metrics} between our approach and the baseline methods under two settings. The two metrics reflects different aspect of the joint matching quality.
		Joint chamfer distance evaluates the average distance between matched peg-hole, but does not reflect whether joints are successfully aligned. 
		Joint matching accuracy evaluates the number of aligned peg-hole pairs among all peg-hole joints for the shape. 
		The bottom rows indicates that our joint input and loss setting helps baselines to lower the distance between all joints, but resulting in collapsed shapes and thus worse matching accuracy.}}
		\label{table:exp-joint}
        \vspace{-2.5mm}
	\end{table*}	

\textbf{Joint loss}
The joint matching task is very sensitive to prediction errors; one small matching error can lead to the deterioration of the entire shape. Therefore, we supervise for the joint matching objective in a coarse-to-fine manner with three loss components:
$\mathcal{L}_{joint} = \lambda_4 \mathcal{L}_{flip}  + \lambda_5 \mathcal{L}_{coarse} + \lambda_6 \mathcal{L}_{fine}$. 
The first loss term $\mathcal{L}_{flip}$ directly corrects the flipped pose predictions. 
Inspired by~\cite{li2020learning}, we use rotation L2 loss to correct upside-down predictions for parts with reflective symmetry:
\begin{equation}
    \mathcal{L}_{flip} = \sum_{i=1}^N \left\lVert q_i(p_i) -  q_i^{gt}(p_i)\right\rVert_F^2,
\end{equation}
The second loss term $\mathcal{L}_{coarse}$ provides coarse guidance to attach the matched joints. We use L2 distance between the matched pegs $j_a^{p}$ and holes $j_b^{h}$. We use $n_{joint}$ denotes the number of joint points,
\begin{align}
\mathcal{L}_{coarse} = & \sum_{{i} = 1}^{M} || q_a(\bar{j_a^{p}}) -  q_b(\bar{j_b^{h}}) ||_2^2, \quad \bar{j_a^{p}}=\frac{1}{n_{joint}} j_a^{p}
\label{eq:jointl2}
\end{align}
The last loss component $\mathcal{L}_{fine}$ uses joint geometric cues to refine joint alignment, inspired by previous works~\cite{zhang20153d, huang2006reassembling}. We use Chamfer Distance between the paired peg $j_a^{p}$ and hole $j_b^{h}$ with predicted poses applied,
\begin{equation}
\begin{small}
  \begin{aligned}
   \mathcal{L}_{fine} = d_{chamfer}(q_a(j_a^{p}), q_b(j_b^{h})),  \quad j_{a}^p \in p_a,  j_{b}^h \in p_{b}
\label{eq:jointcd}
\end{aligned}
\end{small}
\end{equation}
The latter two components of the joint losses are conditioned on a joint matching assignment $\Phi: \{\phi_{i}=(j_{a}^p,j_{b}^h)\}$. Since any arbitrary permutations in the congruent part class are also valid predictions, we cannot directly use the ground truth joint matching $\Phi_{gt}$ as our supervision signal. Therefore, to guarantee the order invariance of joint matching, we design a joint-matching algorithm with a graph traversal scheme to reassign matching of joints between congruent part classes (details in Alg.~\ref{alg:matching} in Supplementary Material). 
\label{sec:matching}

\textit{Remark. }
To tackle the intertwined problem of joint-centric part assembly, we propose an iterative hierarchical graph learning approach. We use two subgraph embeddings to focus on different aspects of the bilateral objectives. The part graph learns to predict and refine part poses to optimize the shape structure. The joint graph messaging passing discovers the joint-wise relationships. The two kinds of learned messages are combined to predict part poses to construct shapes that are both structurally sound and joint aligned. 
\section{Experiments}
\label{sec:exp}
We introduce a joint-augmented part dataset as well as a set of evaluation metrics to examine both the shape-structure and joint-alignment aspects of task performance. 
We compare with six re-purposed prior works to demonstrate that our proposed method is more effective. We conduct ablation studies to validate our design choices.
\subsection{Dataset}

	We adapt the PartNet \cite{mo2019partnet} for our task by augmenting the parts of the shapes with joint annotations. Following \cite{li2020learning}, we use the three largest furniture categories that require real-world assembly, chairs, tables, and cabinets, and adopt the PartNet official train/validation/test split. We use Furthest Point Sampling (FPS) to sample 1,000 points over each part mesh. All parts are canonicalized to be zero-centered and rotated to be local axis aligned using PCA. We detect joint points by computing all pair-wise part Chamfer Distance (eq.~\ref{equation:chamfer}) and take the closest 50 points between two connected parts with a minimum distance of less than 0.05. Following~\cite{li2020learning, qingnan}, we use Level-3 granularity and filter out the shapes with more than 50 pairs of joints, which leaves us with 3736 chairs, 5053 tables, 719 cabinets.

\begin{figure*}
    \centering
    \includegraphics[width=1\linewidth]{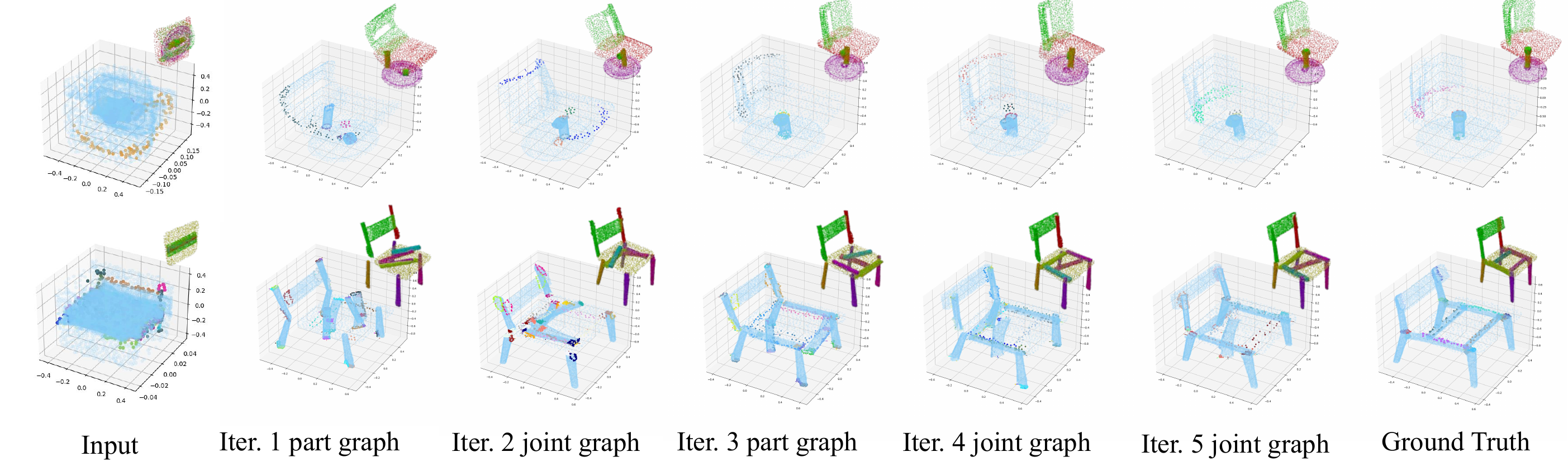}
    
        \caption{\small{Examples of coarse-to-fine part pose prediction over different iterations of our network. The iterations are interleaved graph convolutions of part graph and joint graph. }}
    \label{fig:graph_stage}
        \vspace{-2.5mm}
\end{figure*}
	
\subsection{Baseline Methods}
\label{sec:baseline}

	Since our task is novel, there is no direct comparison from previous works that address the exact joint-alignment aspect of multi-part shape assembly. Instead, we re-purpose previous works, which are originally proposed for part-based shape modeling, to make six baseline methods, as described below (implementation detail can be found in Appendix~\ref{supp:detail}).

    \noindent \textbf{B-Complement:} Sung et al.~\cite{sung2017complementme} propose to address the task of shape generation by retrieving part candidates from a large part repository. Following \cite{qingnan, li2020learning}, we modify the application setting to assemble parts in a sequential manner.

	\noindent \textbf{B-LSTM:} Inspired by the sequential part generation work, such as PQ-Net~\cite{wu2019pq, harish2021rgl}, B-LSTM utilizes an LSTM backbone structure to sequentially decode part poses conditioned on previous part pose estimation. 

	\noindent \textbf{B-Global:} Inspired by CompoNet \cite{schor2019componet} and PAGENet \cite{Li:2020}, B-Global augments part attributes with the global context when decoding part poses. B-Global is commonly used as a baseline method in existing part assembly works~\cite{li2020learning, qingnan, harish2021rgl}.

    \noindent \textbf{B-GNN:} Previous works~\cite{qingnan, harish2021rgl} propose to use an iterative graph neural network to assemble a variety of shapes given a part set. B-GNN follows \cite{qingnan, harish2021rgl} to use dynamic graph learning for our joint-centric part assembly task.
    
\noindent \textbf{B-NSM:} Chen et al.~\cite{NSM} propose a two-part mating network by regressing part poses using a transformer and adversarial training. B-NSM adapts this method to our multi-part multi-joint setting, using a transformer with self-attention.  
    
\noindent \textbf{B-Joinable:} Willis et al.~\cite{willis2022joinable} tackle the task of joint-centric assembly of \textit{two} watertight volumetric parts by joint axis prediction. B-Joinable adapts this method to our setting. 
\subsection{Evaluation Metric}
\label{sec:metric}
\textit{Shape Evaluation Metric. } Following \cite{li2020learning}, we adopt the two metrics of part accuracy \textit{(Part Acc.)} and shape chamfer distance \textit{(Shape CD)} to evaluate the assembled shape structure. Part accuracy threshold $\tau_p$ is chosen to be 0.1.
\begin{equation}
\begin{aligned}
\text{\textit{Part Acc.}} &= \frac{1}{N} \sum_{i=1}^N \mathbb{1}\left[ d_{chamfer}\left( q_i(p_i), q_{i}^{gt}(p_i)\right)\right] < \tau_p
\end{aligned}
\end{equation}
\begin{equation}
\begin{aligned}
\text{\textit{shape CD}} &= \frac{1}{N \cdot n_{points}} \sum_{i=1}^N d_{chamfer} \left(q_i(p_i), q_i^{gt}(p_i)\right)
\end{aligned}
\end{equation}

\textit{Joint Evaluation Metric. }  We propose two metrics for evaluating the second objective of the joint-centric part assembly task, joint accuracy \textit{(Joint Acc.)} and joint chamfer distance \textit{(Joint CD)}. Specifically, \textit{Joint Acc.} evaluates how many joints are matched under the order-invariant joint matching algorithm defined by Alg.~\ref{alg_joint_matching} (in Supplementary Material). It measures the percentage of joint pairs with Chamfer Distance under chosen threshold $\tau_j$=0.01.  
\begin{equation}
\begin{aligned}
\text{\textit{Joint Acc.}} = \frac{1}{M} \sum_{\phi_{i} = 1}^{M} \mathbb{1} \left[ d_{chamfer}(q_b(j_b^{h}), q_a(j_a^{p})) \right] < \tau_j
\end{aligned}
\end{equation}
where $\phi_i=(j_{a}^p,j_{b}^h)$ and $j_{a}^p \in p_a,  j_{b}^h \in p_{b}$.

Additionally, we also resort to joint Chamfer Distance metric to reflect the preciseness of part joint alignment by each method,
\begin{equation}
\begin{aligned}
\text{\textit{Joint CD}} &= \frac{1}{M}\sum_{\phi_{i} = 1}^{M} d_{chamfer}\left( q_a(j_a^{p}), q_b(j_b^{h})\right)
\end{aligned}
\end{equation}

\subsection{Results and Analysis}

  To guarantee the fairness of comparison, we devise two settings for the baseline experiments: (\textit{I}) the setting in their original formulation, as also adopted in \cite{qingnan, li2020learning, harish2021rgl}; (\textit{II}) baseline models trained with the same joint inputs and losses used in our model using the same loss weights as our model. The original task formulation of B-Joinable~\cite{willis2022joinable} explicitly considers joints, so we show its results in setting II.

\textit{Setting I. } By explicitly modeling joint connections, our method improves shape structure over previous methods that purely consider part geometries. 
Table~\ref{table:exp-shape} compares our proposed methods and baseline methods according to the shape structure metric. 
We can see from the table that our method consistently outperforms baseline methods on the shape metric.
Qualitative evidence in Fig.~\ref{fig:qual} (shape structure in the top-right corner) also shows that our predictions are structurally similar to the ground truth, outperforming all the baselines. 
We can observe from the joint-centric view of the figure (blue shapes joint pairs are in the same color) that the baseline methods make flipped or upside-down predictions for parts with rotational and reflective symmetry, e.g. the chair armrest and seat poses predicted by B-GNN. Additionally, in their original proposed setting, the baseline methods only consider part geometries and can be confused by parts with similar geometries. Thus, they cannot determine the correct pose for these parts. For example, Fig.~\ref{fig:qual} shows, cabinets are made with boards of similar shapes. B-LSTM and B-Complement place all boards horizontally.
Joints provide additional information for the functionality of each part. The two vertical side boards have multiple parallel joints, and the horizontal racks have joints at the two ends. Our method utilizes this information to achieve better shape structures.

\textit{Setting II. } 
The performance of the baseline methods degrades significantly under experimental setting II, compared with setting I. 
We can tell from Table~\ref{table:exp-shape} that the bottom section showing setting II is almost one-third of the performance of \textit{setting I}, their original proposed setting, shown on the top section of the table. This demonstrates that our multi-part multi-joint task is challenging. Baseline methods cannot be directly adopted to tackle our task that involves two different objectives with two kinds of relationship reasoning.
Similar phenomena can be observed in Fig.~\ref{fig:qual}, in which the bottom three rows show the baseline predictions under setting II. We can see that the baselines predict collapsed shapes. This is because when all parts are clustered together, their joint distances are also decreased. This type of collapsed prediction is a local minimum for the joint losses.

This observation gives us insights into the loss landscape of the two task objectives. The shape structure objective aims to spread out the part geometries to various locations. 
The joint alignment objective aims to connect and contact parts together. For any inaccurate pose predictions, these two objectives conflict with each other. Shape structure wants to expand parts, whereas joint matching wants to contract parts. There exists only one set of pose predictions that simultaneously satisfy both task objectives--that is, the global optimum.
This explains why most baseline methods fail when they are subjected to both objectives at the same time. The local minimum for the joint matching objective can easily trap pose predictions for baseline methods from achieving valid shape structures. However, our method maintains the valid shape structure using a coarse-to-fine scheme, and thus is less prone to stuck in the local minimum of collapsed shape.

\begin{table}[]
    \centering
    \begin{small}{
\begin{tabular}{c|c|c|c|c}
	\toprule
    				& SCD $\downarrow$ & PA $\uparrow$ & JCD $\downarrow$ & JA $\uparrow$ \\ 

    				  \midrule				
    				w/o 1st iter of part graph & 0.025 & 23.71 & 0.572 & 33.86 \\

    				w/o joint embedding & 0.019 & 33.36 & 0.258 & 36.90  \\
    				w/o matching alg.& 0.010 & 55.10 & 0.416 & 37.50   \\

    				\midrule				
    				Ours full & 0.006 & 72.81 & 0.352 & 57.18     \\
    		
    				\bottomrule
    			\end{tabular}
			\caption{\small{Ablation study conducted on the Chair category. SCD denotes Shape Chamfer Distance; PC denotes Part Accuracy; JCD denotes the Joint Chamfer Distance, and JA denotes Joint Accuracy. Arrows indicates the direction of better performance.}}
			\label{table:ablation}
    }\end{small}
    \vspace{-3mm}
\end{table}
 
In Figure~\ref{fig:graph_stage}, we show our coarse-to-fine assembly scheme by visualizing the predicted shape structures in the intermediate stages of our graph convolution. We observe that the first iteration of part graph learns to predict a rough shape structure. The subsequent joint graph iteration learns to modify part poses so more parts can be connected with each other. Another iteration of part graph then learns to refine part poses with the corrected joint matching. Eventually, through these iterations of graph convolution, we can produce structurally sound and joint-matched part assemblies. We use the part-joint-part-joint-joint graph combination, as we discovered that this combination works best empirically.

 \textit{Ablation Study. } We conduct three ablation experiments to demonstrate the effectiveness of different design choices of our proposed approach, as shown in Table~\ref{table:ablation}. 
We first test our network design of using a part graph module as the first iteration of our network. We believe that the rough shape structure proposed by the first part graph module can serve as a shape structure heuristic to reduce the joint matching difficulty. As shown on the top row in Table.~\ref{table:ablation}. Removing the first iteration of the part graph by directly having the joint graph to propose joint matching solutions significantly reduce our performance, and hence verifies our conjecture of shape structure heuristic on joint matching. 

Our second ablation experiment aims to test the importance of the joint embedding, as shown on the second row in Table.~\ref{table:ablation}. We remove the joint embedding step and applying the joint losses on the last stage part graph directly. The result shows a significant performance decrease in \textit{Shape CD} and \textit{Part Accu.}. This setting is similar to experimental setting II by directly adding joint losses to baseline methods. This shows that joint embedding is a non-trivial component to our network that provides more explicit joint-alignment pose editing signal.

We then test our method without the matching algorithm in the loss scheme, as described in Alg.~\ref{alg_joint_matching} in Appendix. We observe that the performance decreases significantly across all metrics. This is because Hungarian-matching allows shape losses to be permutation invariant for geometrically congruent parts, the joint matching algorithm maintains this order-invariance property for the joint losses. Alg.~\ref{alg_joint_matching} finds a new matching assignment considering permutations within the congruent part class, granting consistency between the two loss objectives. Without such consistency, the two losses are not synchronized and work in different directions, and thus result in problematic part pose predictions.

\section{Conclusion and Future Work}
\label{sec:conclusion}

We formulate a novel variant of the category-level multi-part 3D shape assembly problem by introducing the concept of joints. We focus on the peg-hole abstraction of part joints and proposed a hierarchical graph network approach that consists of a joint embedding and a part embedding for explicit hierarchical relationship reasoning to tackle the challenges. We introduce a joint-augmented multi-part assembly dataset along with evaluation metrics to set up the test bed for this task. We also provide extensive empirical evidence to demonstrate the effectiveness of our approach compared to the re-purposed prior works. We believe our work can  on autonomous assembly systems.

As a start of the multi-part multi-joint assembly problem, we focus on the simple but commonly-used peg-hole joints. There are several possible scenarios that are not considered in our paper and are left for future work. One future direction is to extend to more complicated joints for this problem and construct a general formulation for all possible joint types. Another future direction is programmatic or sequential planning for joint alignment, which would better enable vision algorithms to be deployed in autonomous systems. 

\clearpage

{\small
\bibliographystyle{ieee_fullname}
\bibliography{refs.bib}
}

\clearpage
\newpage
\section{Appendix}
\renewcommand{\thefootnote}{\arabic{footnote}}
Supplementary material includes
\begin{itemize}
    \item {Order Invariant Joint Matching Algorithm}
    \item Implementation Detail
    \item Additional Qualitative results.
    \item Additional Numerical Comparisons.
\end{itemize}

\subsection{Order Invariant Joint Matching Algorithm}
\begin{algorithm}[h]			
	\caption{Order Invariant Joint Matching Algorithm} \label{alg:matching}
	\begin{algorithmic}[1]
		\State Sort all the nodes in the connectivity graph by the degree of the target node in descending order.
		\While{not all the nodes are assigned}
		    \State Start with the most connective node as the target node \textit{i} and assign it with sign \textit{peg}.
		    \ForAll{neighbor node \textit{j} of the target node}
		        \If {\textit{j} has been assigned with \textit{peg}}
		            \State Add (\textit{i}, \textit{j}) pair to conflict cache
		        \Else
		            \State Assign \textit{hole} to neighbor node \textit{j}
		        \EndIf

            \EndFor
		\EndWhile
		\ForAll{edges \textit{e} in the connectivity graph}
		    \If{the two nodes of \textit{e} have the same label}
		        \State Add the edge to conflict cache
		    \EndIf
		\EndFor
        \ForAll{edge=(\textit{i},\textit{j}) in conflict cache}
            \If{one of \{\textit{i}, \textit{j}\} is congruent}
                \State Assign congruent part with \textit{peg} 
            \Else
                \State Remove (\textit{i}, \textit{j}) from connectivity graph
            \EndIf
        \EndFor
    \end{algorithmic}
    \label{alg_joint_matching}
\end{algorithm}

\begin{figure}[h]
    \centering
    \vspace{-2mm}
    \includegraphics[width=1\linewidth]{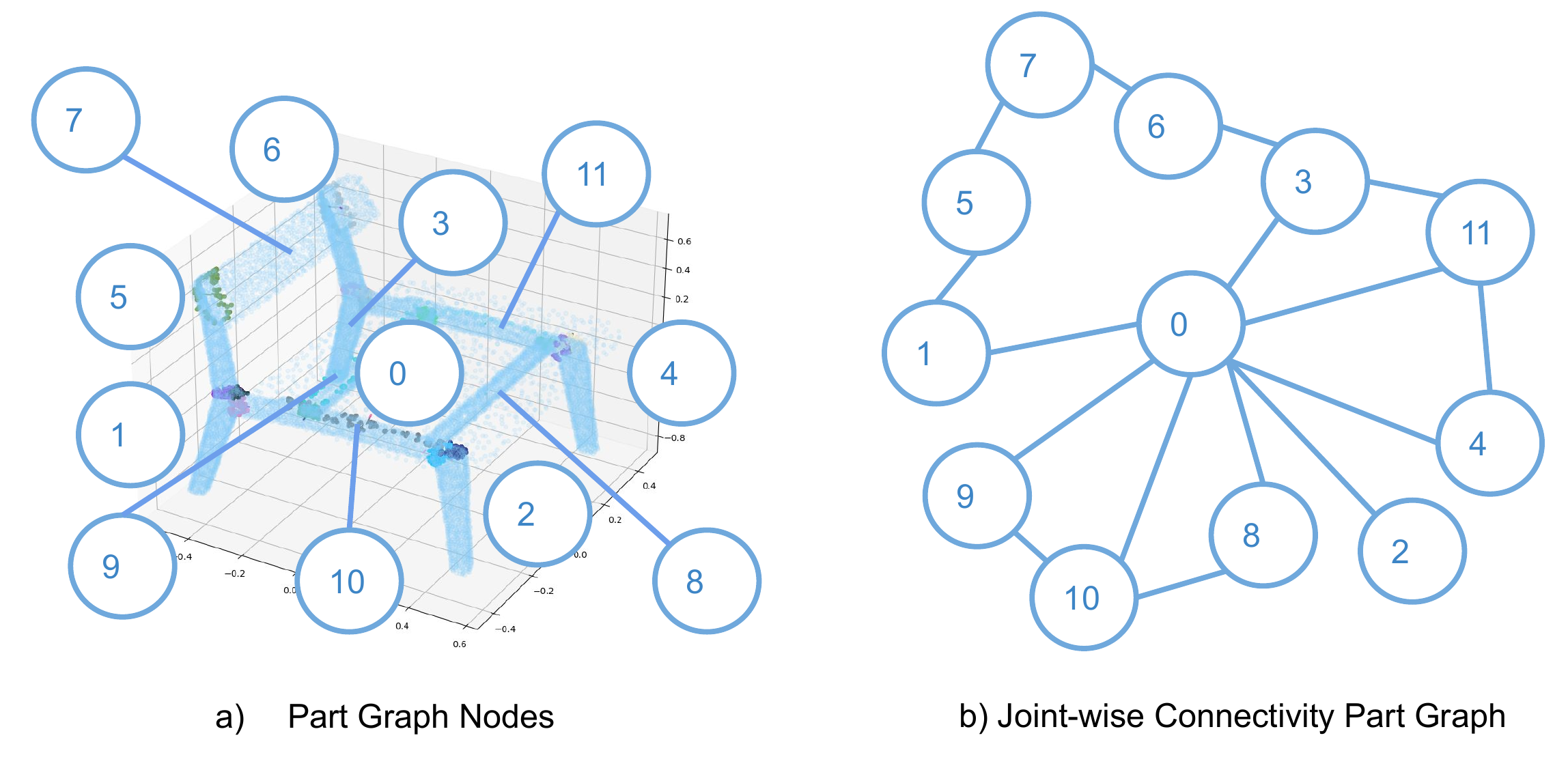}
    
        \caption{{Example joint-wise connectivity graph for a ShapeNet Chair}}
    \label{fig:supp-matching-detail}
    \vspace{-2mm}
\end{figure}

            
The permutation invariance nature of the losses schemes forbids us to use the ground truth matched pairs, as there lack of a global rule in the ground truth joint matching annotation that defines what should be the peg and what should be the hole. For example, for four chair legs that are connected to the chair seat, a set of the leg joints are defined as pegs and the other ones are defined as holes and vice versa for their mated seat joint. In Figure~\ref{fig:supp-matching-detail}, the four legs are geometrically interchangeable. However, the ground truth joint matching annotation defines the front right leg, part 2, and its corresponding seat joint are connected via leg-peg and seat-hole joint pair. The front left leg, part 4, and its corresponding joint are connected via leg-hole seat-peg joint pair. This ground truth matching scheme prevents the chair legs to be interchangeable if the network predicts the reversed placement for the two legs, as the matching rule defines that only different types of joints can be mated. For example, if part 2 is now in place of part 4, which means that the leg-peg and seat-peg are now placed next to each other.

A second problem is with part joints that are very close to each other on the same part for example the back legs of the example chair in Figure.~\ref{fig:supp-matching-detail} have three joints in the same area, and all of them need to be the same type, so that there is a signal that tells the network that they cannot be joined because otherwise these joints that are always very close to each other regardless of the part pose prediction. Yet, simply defining joints from the same part to be of the same sex does not do the job, as it will fail when there are loops in the joint graph, especially with odd element loops. For example, part 0, part 9, and part 10. If part 0 is defined to be a peg part, then part 9 and part 10 will simultaneously become hole parts. However, parts 9 and part 10 are connected to each other, thus this simple scheme does not work in such cases. We introduce the Order-Invariant Joint Matching Algorithm that first leverages graph traversal to assign joint-type in a part-wise manner and then deals with the conflicted assignment separately, as shown in Alg.~\ref{alg_joint_matching}.

\subsection{Additional Explanation of the Task}

Here, we provide an additional explanation of our proposed category-level multi-part multi-joint assembly task. The main difference between the proposed task and the task of shape generation from part assembly lies in the joint-matching constraint. Without the joint-matching constraint, many different shape geometries can be composed of the given set of parts. For example, the rectangular chair seat can be rotated in different ways and still compose a valid chair design. However, joint-matching is a strong constraint. For example, the upside-down prediction of the same chair seat is no longer valid because such a pose will not be able to connect with the chair legs. In a less obvious case, a prediction rotated by 90 degrees also may not be able to satisfy the constraints as the chair back will not be able to properly attach to the chair seat. As shape structure becomes more complex, part poses that satisfy the joint-matching objective become more constrained. 

We noticed empirically the joint-matching objective eliminates most shapes of geometric variations in the variational generation task and only the exchange of congruent parts are preserved. Therefore, we frame the multi-part multi-joint part assembly problem as a single-solution problem.

\subsection{Implementation Detail}
\label{supp:detail}
\begin{figure}[]
    \hspace{-1mm}
    \vspace{-0.2cm}
    \begin{center}
    \includegraphics[width=0.7
\linewidth]{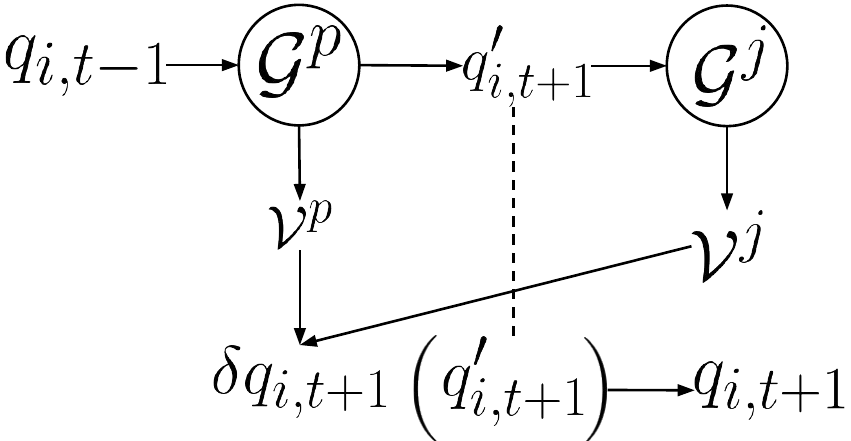}
    \end{center}
      \caption{Overview of one graph iteration for pose prediction, composed of the part graph and the joint graph. }
    \label{fig:supp-method}  
    \vspace{-0.5cm}
\end{figure}

\textit{Our Method. } Here, we offer some additional explanations of the technical and implementation detail of our method. Figure~\ref{fig:supp-method}, shows one iteration pose prediction. For the first iteration, $q_{i, 0}$ is initialized to be 0 for translation (0, 0, 0) and 0 for rotation (quaternion [0, 0, 0, 1]).

\textit{Part Graph} The input to our part graph is a set of part point clouds, each with 1000 points denoting the $x,y,z$ coordinates in $\mathbb{R}^3$. We use PointNet~\cite{qi2017pointnet} to extract the geometric feature for each part to be used as a node feature for the graph. We conduct message passing between all part nodes by iteratively updating node and edge features $f_{v \rightarrow e}, f_{e \rightarrow v}$. Finally, we predict part pose $q_i$ for each part $p_i$ from the updated node feature for that part $v_i$ using MLP. Each part pose $q_i \in \mathbb{R}^{7}$ is composed of a unit 4-dimensional quaternion rotation vector and a 3-dimensional translation vector.

\textit{Joint Graph} The input to our joint graph is a joint-masked four-dimensional point cloud. Speficially, we organize our input joint peg-hole points $\mathcal{J} = \{j_i^{p,h}\}_{i=1}^M$ as ternary mask $\{m_i^{p,h}\}_{i=1}^M, m_i^{p,h} \in \{-1,0,+1\}$ for the part point clouds $\mathcal{P} = \{p_i\}_{i=1}^N$, where the mask values of $-1$ denote the peg points, $+1$ for hole points, and $0$ for non-joint point. Applying each joint mask to its parent part point cloud $\mathcal{J}^{p}=\{p^{joint}_{i}=(p_i; m_i^{p,h})\}_{i=1}^{2M}$ forms 4-dimensional point cloud containing information of both joint points and geometry of the part $\mathcal{P}^{j}=\cup_{t=1}^{n_{i}}\{p^{joint}_t\}$. We use a 4-dimensional PointNet to extract the joint geometric feature for each joint point cloud to form the joint embedding in the joint graph. Then we conduct message-passing on the joint graph. 
All message-passing modules $f_{v \xrightarrow e}, f_{e \xrightarrow v}, $, relationship reasoning modules$f_{v \xrightarrow r}, f_{r \xrightarrow v},$and pose prediction modules $f_{pose}$ are modled using Multi-Layer Perceptron (MLP) with hidden dimension of 128.

\textit{Loss Function Parameters} Following~\cite{li2020learning}, we define $\mathcal{L}_{shape}$ as a weighted combination of local part translation loss $\mathcal{L}_t$, rotation loss $\mathcal{L}_r$ and the global shape structure loss $\mathcal{L}_s$. Empirical evidence has shown that parameter values of $\{\lambda_1=1.0, \lambda_2=10.0, \lambda_3=1.0\}$ produce the best results. We define our joint loss $\mathcal{L}_{joint}$ to be a weighted combination of L2 distance between the means of the matched joints $\mathcal{L}_{jL2}$, Chamfer distance of matched joint points $\mathcal{L}_{jcd}$ as well as local part L2 rotation loss $\mathcal{L}_{rL2}$. Empirical evidence has shown that loss weights of $\{\lambda_4=1.0, \lambda_5=5.0, \lambda_6=1.0\}$ produce the best results. All baselines in setting II are conducted using the same loss weights of $\{\lambda_1=1.0, \lambda_2=10.0, \lambda_3=1.0, \lambda_4=1.0, \lambda_5=5.0, \lambda_6=1.0\}$.

\textit{Baselines. } We follow the same implementation scheme used in \cite{qingnan, harish2021rgl} for B-Complement, B-LSTM, B-Global, B-GNN, using the official code release provided by Huang et al.~\cite{qingnan}. 
For B-NSM, we use the official code release by Chen et al.~\cite{NSM}, and we modify the attention module in the transformer. As the original setting only considers two parts and the original attention module used in \cite{NSM} compares the feature vectors of the two parts. Since we tackle a multi-part multi-joint setting, where different shapes contain a various number of parts and different parts contain a various number of joints. Directly applying pair-wise attention on all possible pairs between all parts and all joints is computationally inefficient. Therefore, we encourage implicit relationship reasoning using self-attention by applying the same attention module on feature vectors compressing all parts and joint information of a given part set. 

B-Joinable modifies the original method~\cite{willis2022joinable} to take point cloud data of multiple parts. We preserve the geometry encoding and joint-axis prediction philosophy of the original method. We encode part point cloud data and predict a 3 DOF axis for each matched joint. The predicted joint axis is used to further predict the 6 DOF poses for each part. The contact B-Rep joint loss functions proposed by Willis et al. have similar objectives to our joint losses for point clouds.
In addition to the setting of imposing our loss functions, we also experimented with supervising for joint axis.
We add additional supervision signals calculated from contact joints normals. Empirically, we discover this additional supervision signal hurts performance. Because without the correct pairing of joints, the joint axis signal adds noise to the supervision. In our paper, we report the best-performing scenario for our task. 


\textit{Hardware for Training. } We train and test all baselines and our method using a workstation equipped with a 16-core AMD CPU, and 2 NVIDIA GTX 3090 GPUs. All methods are trained to full convergence for 18 hours.  

\subsection{Additional Qualitative Results}
\begin{figure*}
    \includegraphics[width=1\linewidth]{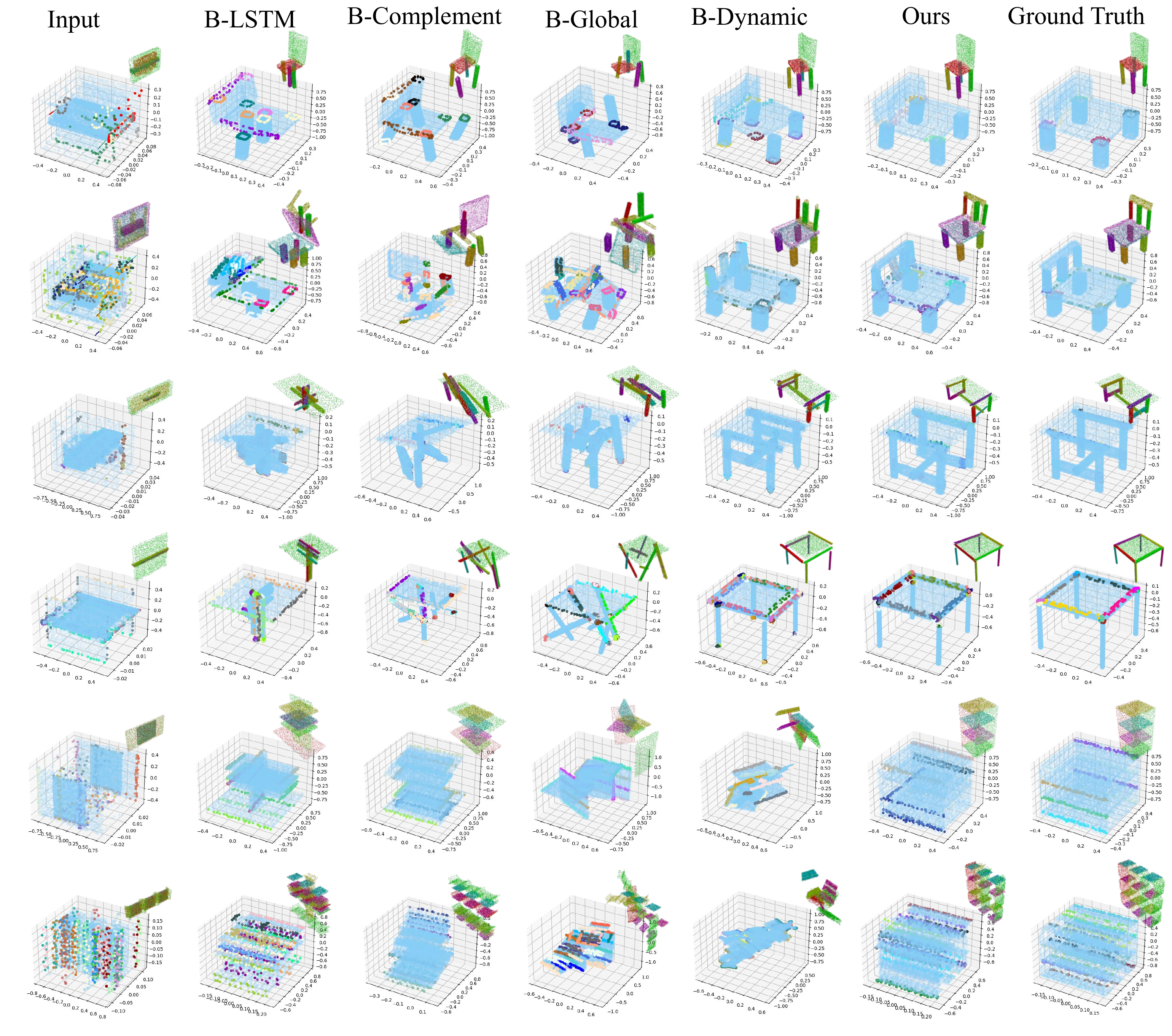}
      \caption{{Qualitative comparison of our method and baselines in their best performing setting, their original setting.}}
    \label{fig:supp-qual}  
\end{figure*}

Additional qualitative examples are shown in Figure~\ref{fig:supp-qual}.

\subsection{Additional Baseline Comparisons }
In addition to the qualitative results shown in Section~\ref{sec:exp}, we also provide the baselines' performance with our joint-aware part input but use their original losses, shown in Table~\ref{table:exp-supp}. 
We notice that the baselines perform the worst under this setting. We think that adding the joint input does not help the baselines to implicitly infer structured connectivity. Instead, without explicitly joint-centric loss design, the joint information is treated as noise by the baseline methods. This results in decreased baseline performance. Additionally, we also notice that baseline dynamic~\cite{qingnan}  exhibits the lowest performance in the Cabinet category, and cannot infer reasonable pose predictions. 
	\begin{table*}[h]
        \footnotesize
		\begin{center}
		\resizebox{\textwidth}{!}{
	\begin{tabular}{c|c|ccc|ccc|ccc|ccc}
				\toprule
				& & \multicolumn{3}{c}{Shape Chamfer Distance $\downarrow$} & \multicolumn{3}{c}{Part Accuracy $\uparrow$} & \multicolumn{3}{c}{Joint Chamfer Distance $\downarrow$} & \multicolumn{3}{c}{Joint Accuracy $\uparrow$} \\ 
				\midrule
				Setting& Method & Chair & Table & Cabinet & Chair & Table & Cabinet & Chair & Table & Cabinet & Chair & Table & Cabinet \\
				 \midrule				
				 \multirow{4}{*}{Original Setting}&B-Global& 0.015 & 0.013 & 0.008 & 32.8 & 30.1 & 33.6 & 0.712 & 0.847 & 0.667 & 13.4 & 15.8 & 10.7  \\
				&B-LSTM & 0.017 & 0.026 & 0.007 & 39.4 & 22.5 & 44.4 & 0.756 & 0.728 & 0.651 & 17.0 & 13.2 & 14.8  \\
				&B-Complement & 0.028 & 0.034 & 0.222 & 11.0 & 5.33 & 0.0 & 0.901 & 0.977 & 1.074 & 7.54 & 8.30 & 23.6  \\
				&B-Dynamic  & 0.007 & 0.008 & 0.006 & 65.3 & 61.4 & 45.0 & 0.725 & 0.855 & 0.683 & 24.4 & 30.0 & 18.6    \\
				\midrule
				\multirow{4}{*}{\shortstack{Our Input Setting \\ (joint input)}}&B-Global & 0.076 & 0.031 & 0.028 & 3.9 & 12.4 & 5.7 & 1.724 & 1.239 & 1.281 & 3.2 & 4.8 & 2.7 \\
				& B-LSTM &  0.028 & 0.039 & 0.010 & 7.3 & 3.8 & 27.2 & 0.777 & 0.708 & 0.661 & 8.7 & 15.1 & 6.7    \\
				&B-Complement & 0.055 & 0.047 & 0.015 & 2.9 & 7.2 & 15.8 & 1.413 & 1.004 & 0.767 & 5.9 & 6.5 & 10.1    \\
				&B-Dynamic  &  0.046 & 0.076 & 0.035 & 4.3 & 2.1 & 3.8 & 0.547 & 1.120 & 0.639 & 15.4 & 8.1 & 7.8 \\
				\midrule
				\multirow{4}{*}{\shortstack{Our Full Setting \\ (joint input and loss)}}&B-Global & 0.029 & 0.022 & 0.013 & 5.4 & 12.0 & 15.0 & 0.513 & 1.268 & 0.488 & 12.7 & 4.0 & 6.9  \\
				& B-LSTM & 0.037 & 0.029 & 0.017 & 4.4 & 4.1 & 15.3 & 0.394 & 0.875 & 0.467 & 20.3 & 7.7 & 13.8   \\
				&B-Complement &0.048 & 0.044 & 0.029 & 4.5 & 8.0 & 11.6 & 0.456 & 0.647 & 0.503 & 17.2 & 15.5 & 17.0   \\
				&B-Dynamic  & 0.034 & 0.039 & 0.021 & 11.5 & 3.2 & 10.4 & 0.379 & 0.786 & \textbf{0.416} & 21.5 & 10.3 & 20.0  \\
			    \midrule
				&\textbf{Ours} & \textbf{0.006} & \textbf{0.007} &\textbf{ 0.005} & \textbf{72.8} & \textbf{67.4} & \textbf{63.3} & \textbf{0.352} & \textbf{0.602} & 0.620 & \textbf{57.2} & \textbf{50.6} & \textbf{27.5}  \\
				\bottomrule
			\end{tabular}}
		\end{center}
		\caption{Quantitative comparison between our approach and the baseline methods.}
		\label{table:exp-supp}
	\end{table*}	



\end{document}